\documentclass[times,twocolumn,final,authoryear]{elsarticle}

\usepackage{ycviu}
\usepackage{framed,multirow}
\usepackage{amsmath,amssymb}%
\usepackage{latexsym}
\usepackage[linesnumbered,vlined,ruled]{algorithm2e}

\usepackage{url}
\definecolor{newcolor}{rgb}{.8,.349,.1}
\usepackage[dvipsnames]{xcolor}
\usepackage{xcolor}

\newcommand{\ie}{\textit{i}.\textit{e}.}
\newcommand{\eg}{\textit{e}.\textit{g}.}
\usepackage{pifont}

\journal{Preprint}
\begin{document}
\thispagestyle{empty}

\clearpage

\setcounter{table}{0}

\ifpreprint
  \setcounter{page}{1}
\else
  \setcounter{page}{1}
\fi

\begin{frontmatter}

\title{Reverse Stable Diffusion: What prompt was used to generate this image?}

\author[1]{Florinel-Alin \snm{Croitoru}}
\author[1]{Vlad \snm{Hondru}} 
\author[1]{Radu Tudor \snm{Ionescu}\corref{cor1}}
\cortext[cor1]{Corresponding author at: Department of Computer Science, University of Bucharest, 14 Academiei Street, Bucharest 010014, Romania}
\ead{raducu.ionescu@gmail.com}
\author[2]{Mubarak \snm{Shah}}

\address[1]{Department of Computer Science, University of Bucharest, 14 Academiei Street, Bucharest 010014, Romania}
\address[2]{Center for Research in Computer Vision (CRCV), University of Central Florida, Orlando 32816, FL, US}


\begin{abstract}
Text-to-image diffusion models have recently attracted the interest of many researchers, and inverting the diffusion process can play an important role in better understanding the generative process and how to engineer prompts in order to obtain the desired images. To this end, we study the task of predicting the prompt embedding given an image generated by a generative diffusion model. We consider a series of white-box and black-box models (with and without access to the weights of the diffusion network) to deal with the proposed task. We propose a novel learning framework comprising a joint prompt regression and multi-label vocabulary classification objective that generates improved prompts. To further improve our method, we employ a curriculum learning procedure that promotes the learning of image-prompt pairs with lower labeling noise (\ie~that are better aligned). We conduct experiments on the DiffusionDB data set, predicting text prompts from images generated by Stable Diffusion. In addition, we make an interesting discovery: training a diffusion model on the prompt generation task can make the model generate images that are much better aligned with the input prompts, when the model is directly reused for text-to-image generation. Our code is publicly available for download at \url{https://github.com/CroitoruAlin/Reverse-Stable-Diffusion}.
\end{abstract}


\end{frontmatter}

\section{Introduction}

\begin{figure*}[!t]
\begin{center}
\centerline{\includegraphics[width=0.8\linewidth]{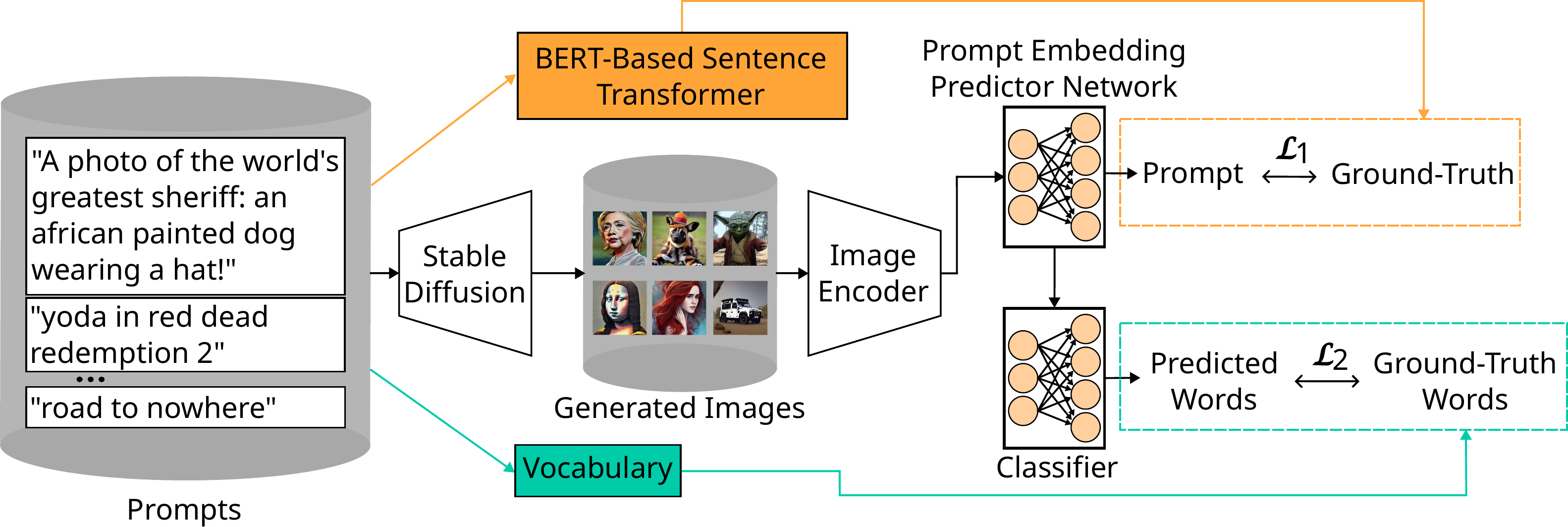}}
\vspace{-0.25cm}
\caption{Our learning framework for prompt embedding estimation, along with its vocabulary classification task. We transform the input prompts via a sentence transformer for the embedding estimation task and we use a vocabulary of the most common words to create the target vectors for the classification task. Best viewed in color.}
\label{fig_pipeline}
\end{center}
\end{figure*}

\begin{figure*}[t]
\begin{center}
\centerline{\includegraphics[width=\linewidth]{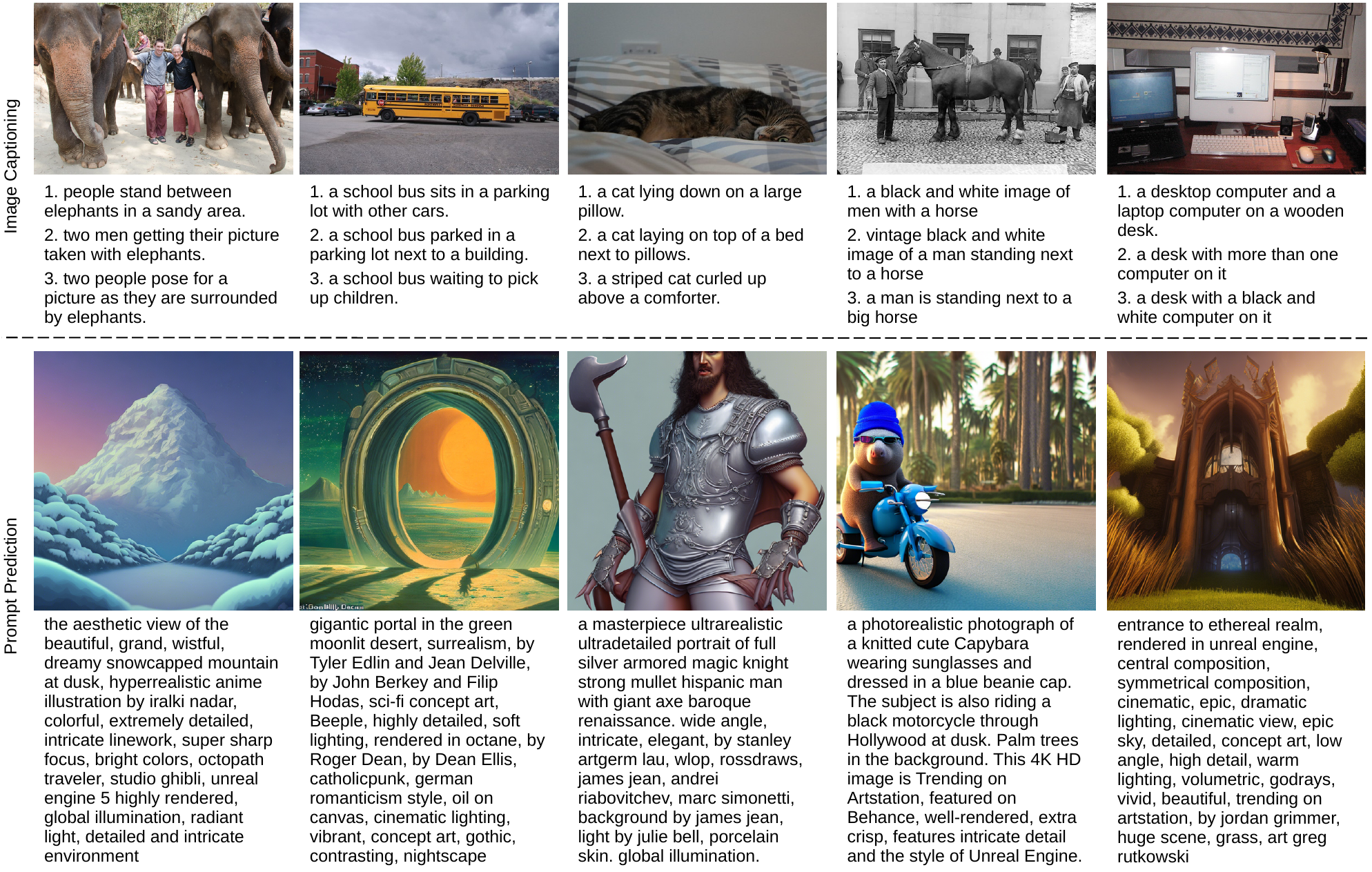}}
\vspace{-0.2cm}
\caption{Comparison between the image captioning and prompt prediction tasks. The samples from the top row are taken from MS COCO, while the samples from the bottom row are taken from DiffusionDB. In prompt prediction, the model must generate a single and very detailed text prompt. In contrast, image captioning benchmarks typically have several alternative ground-truth captions for each image \citep{Chen-Arxiv-2015}, and models are evaluated against the best matching ground-truth caption. Moreover, image captions are generally shorter, referring only to the foreground objects and their interactions. Best viewed in color.}
\label{fig_prompt_vs_caption}
\end{center}
\end{figure*}
Diffusion models \citep{Croitoru-PAMI-2023,song-ICLR-2021, ho-NeurIPS-2020, song-NeurIPS-2020, song-NeurIPS-2019, sohl-icml-2015, dhariwal-NeurIPS-2021, nichol-ICML-2021, song-ICLR-2021} have recently emerged as a powerful type of deep generative models. They are used in various tasks, such as image generation \citep{song-ICLR-2021, ho-NeurIPS-2020, song-NeurIPS-2020}, super-resolution \citep{saharia-arXiv-2021, daniels-NeurIPS-2021, rombach-CVPR-2022, chung-CVPR-2022}, image editing \citep{avrahami-CVPR-2022}, text-to-image generation \citep{rombach-CVPR-2022, ramesh-arXiv-2022, saharia-arXiv-2022}, and many others \citep{Croitoru-PAMI-2023}. Text-to-image generation has gained significant attention, with models such as Stable Diffusion \citep{rombach-CVPR-2022} standing out for their efficiency and effectiveness, using the diffusion process in a compact semantic space for rapid and high-quality image synthesis conditioned on text.
Despite the positive results in text-to-image generation, there is a notable lack of research regarding the understanding of these models. For example, there is a rising need to understand how to design effective prompts that produce the desired outcome. 
Further investigation in this area will be crucial for the advancement of text-to-image generation models. For instance, although the generated images are usually correlated with the input prompts, there are many situations when certain tokens from the text prompt are completely ignored, or some latent properties of the respective tokens are not included.

To this end, our work focuses on studying the 
task of reversing the text-to-image diffusion process. Due to its notoriety, we particularly focus on reversing the Stable Diffusion model. Given an image generated by Stable Diffusion, the proposed task is to predict a sentence \textit{embedding} of the original prompt used to generate the input image. Although predicting the actual prompt is also possible, we consider that this form of the reverse task is ill-posed. As illustrated in Figure~\ref{fig_prompt_vs_caption}, the primary difference between prompt embedding prediction and image captioning is at the semantic level. Captions are typically short sentences that refer to the foreground objects and the interactions between these objects. Moreover, image captioning data sets typically contain alternative captions for the same image \citep{Chen-Arxiv-2015}, such that the captioning model can easily find a match with one of the captions. In contrast, text prompts are typically long sentences, being aimed at accurately describing a mental picture inside the mind of the user. Aside from including the objects and their interactions, prompts typically contain more detailed descriptions of the objects, as well as additional specifications, such as the style of the image, the background, and so on. Moreover, some of the details specified in the prompt might be ignored by the generative model. At the same time, the generative model might also generate certain features that were not specified in the prompt. Therefore, the prompt prediction task is significantly more difficult than image captioning. For the same reason, predicting the exact prompt is more difficult than predicting the prompt embedding. In the embedding space, we essentially allow the models to predict semantically related prompts (paraphrases) without being needlessly penalized. Moreover, this allows us to have a more objective evaluation, avoiding the use of multiple measures to quantify the quality of the generated text prompts, which can be problematic \citep{Callison-Burch-EACL-2006}. Additionally, recent advancements \citep{morris-arxiv-2023} demonstrate that the embeddings generated by state-of-the-art text encoders possess the capability to reconstruct nearly the entirety of the original text. 
Therefore, we assert that the task of predicting the prompt embeddings is as relevant as predicting the original text, and it is more beneficial because it circumvents the intricate challenges involved in assessing the quality of generated text.
Nevertheless, we perform experiments for both prompt generation and prompt embedding prediction. To address the image-to-embedding prediction task, we utilize image encoders to extract feature representations, which are subsequently projected to the sentence embedding space via fully connected layers. As underlying models, we consider three state-of-the-art architectures that are agnostic to the generative mechanism of Stable Diffusion, namely ViT \citep{Dosovitskiy-ICLR-2020}, CLIP \citep{radford-pmlr-2021} and Swin Transformer \citep{Ze-ICCV-2021}. We also include the U-Net model from Stable Diffusion, which operates in the latent space. Notably, we consider both black-box and white-box models (with and without access to the weights of the diffusion network), showing that our novel approach achieves good performance regardless of the underlying architecture. To address the image-to-text generation task, we employ a set of models that are more suitable for captioning, namely BLIP \citep{Li-ICML-2022} and BLIP-2 \citep{Li-Arxiv-2023}.

We propose a novel training pipeline that leads to significant performance gains, regardless of the underlying architecture or the target task. The first novel contribution of our training pipeline is an extra classification head, which learns an additional task, that of detecting the most common words in the vocabulary used to write the training prompts. This classification task constrains our model to produce more accurate text prompts. In Figure \ref{fig_pipeline}, we illustrate this component and our learning method to reverse the text-to-image generative process. Additionally, our second component is an innovative curriculum learning method based on estimating the difficulty of each training sample via the average cosine similarity between generated and ground-truth text prompts, measured over several training epochs. Measuring the cosine similarity over several epochs before employing the actual curriculum gives us a more objective measurement of sample difficulty scores, not influenced by the curriculum or the state of the neural model.
Our third component aims to adapt the model to the target domain via a kernel representation. This component harnesses unlabeled examples from the target domain to compute second-order features representing the similarities between samples from the source (train) and target (test) domains. We call this component the domain-adaptive kernel learning (DAKL) framework, and we apply it as a meta-model on the median embedding of the ViT, CLIP, Swin-T and U-Net models. 

We carry out experiments on the DiffusionDB data set \citep{wang-arXiv-2022}, after filtering out examples with near duplicate prompts. We report comparative results to illustrate the influence of our training pipeline. Moreover, we empirically study the impact of our prompt generation task on the correlation between generated images and input prompts, during text-to-image generation with Stable Diffusion. Notably, our findings reveal that training the diffusion model on the image-to-text prediction task, then reintegrating the model to perform the original task, text-to-image generation, can significantly improve the alignment between the generated images and the input prompts. 



In summary, we make the following contributions:
\begin{itemize}
    \item We thoroughly study the task of image-to-text-embedding prediction to assess the possibility of reversing the text-to-image generation process of diffusion models.
    \item We propose a training pipeline comprising three novel components (a classification head, a curriculum learning method, and a domain-adaptive kernel learning framework) and demonstrate its usefulness on four underlying models.
    \item We showcase a promising application of prompt generation models in text-to-image generation, indicating that a diffusion model trained on generating text prompts can also be used to generate images that are better aligned with their prompts.
\end{itemize}

\section{Related work}


\noindent
\textbf{Image captioning.} One possible approach for the proposed task would be to use an image captioning model to generate a natural language description of an image generated by Stable Diffusion. Next, we can employ a sentence transformer to obtain the prompt embedding. Although our task is semantically and technically different, since the final output is a prompt embedding, we consider image captioning methods \citep{Stefanini-TPAMI-2023} as related work. The earliest deep learning approaches for image captioning \citep{Vinyals-CVPR-2015, Karpathy-CVPR-2015, Mao-ICLR-2015, Fang-CVPR-2015, Jia-ICCV-2015} used CNNs \citep{Hinton-NIPS-2012, Simonyan-ICLR-14, Szegedy-CVPR-2015} as high-level feature extractors, and RNNs \citep{Hochreiter-NeuralComputation-1997} as language models to generate the description conditioned on the extracted visual representations. Further developments on this topic focused on improving both the visual encoder and the language model \citep{Xu-ICML-2015, Lu-2017-CVPR, Dai-ECCV-2018, Chen-CVPR-2018, Wang-CVPR-2017, Gu-AAAI-2018, Xu-ICCV-2019, Li-ICCV-2019, Herdade-NeurIPS-2019, Huang-ICCV-2019, Pan-CVPR-2020}. A major breakthrough was made by \citet{Xu-ICML-2015}, who, for the first time, incorporated an attention mechanism in the captioning pipeline. The mechanism allowed the RNN to focus on certain image regions when generating a word. Thereafter, other attention-based models were proposed by the research community \citep{Lu-2017-CVPR, Huang-ICCV-2019, Pan-CVPR-2020}.

Recent image captioning models have adopted the transformer architecture introduced by~\citet{Vaswani-NIPS-2017}. Most studies \citep{Herdade-NeurIPS-2019, Guo-CVPR-2020, Luo-AAAI-2021} used the original encoder-decoder architecture, followed closely by architectures \citep{Li-ECCV-2020, Zhang-CVPR-2021} inspired by the BERT model \citep{Devlin-NAACL-2019}. Another research direction leading to improvements on the image captioning task is focused on multimodal image and text alignment. This was usually adopted as a pre-training task \citep{radford-pmlr-2021, Li-ICML-2022}. The CLIP (Contrastive Language-Image Pre-training) model introduced by \citet{radford-pmlr-2021} represents a stepping stone in the multimodal field, showing great zero-shot capabilities. It involves jointly training two encoders, one for images and one for text, with the objective of minimizing a contrastive loss, \ie~whether or not the image and the text match. As a result, the cosine similarity between the embedding of an image and that of its description, resulting from the corresponding encoders, should be high, while counterexamples produce a low similarity. \citet{Shen-ICLR-2022} and \citet{Mokady-Arxiv-2021} studied the impact of the CLIP visual encoder on the image captioning task, without additional fine-tuning. The BLIP model \citep{Li-ICML-2022}, short for Bootstrapping Language-Image Pre-training, introduces a novel architectural design and pre-training strategy. Its aim is to create a model that can both understand and generate language.

Our framework is distinct from image captioning approaches because we do not aim to generate natural language image descriptions, but rather to map the images directly into a text embedding space. Moreover, we propose three novel contributions and integrate them into a unified learning framework to significantly boost the performance of the employed models, when these models are used both independently and jointly.

\noindent
\textbf{Diffusion model inversion.} Recent studies, motivated by the application of diffusion models in text-based image editing, have focused on image inversion techniques to retrieve the initial noise that recreates a specified image~\citep{song-ICLR-2021b, Wallace-CVPR-2023, Mokady-CVPR-2023}. In this line of work, \citet{Mokady-CVPR-2023} enhanced DDIM inversion \citep{song-ICLR-2021b} through the optimization of unconditional text embeddings. Meanwhile, EDICT \citep{Wallace-CVPR-2023} leverages coupling layers found in normalizing flows to perform precise image inversions.

Our research aligns more closely with studies \citep{Gal-CVPR-2023, Mahajan-CVPR-2024, Wen-NeurIPS-2023, Ruiz-CVPR-2023} that explore the inversion of the textual embeddings of a diffusion model, with the goal of replicating a given image or a visual concept. For example, DreamBooth~\citep{Ruiz-CVPR-2023} fine-tunes diffusion models to associate specific subjects in images with unique text identifiers. Similarly, \citet{Gal-CVPR-2023} introduced novel vocabulary items into the embedding space of a frozen diffusion model to symbolize concepts depicted through images. Moreover, \citet{Mahajan-CVPR-2024} and \citet{Wen-NeurIPS-2023} introduced per image optimization strategies to create text prompts that yield images similar to a target image.

Distinct from these approaches, our work does not utilize the text embedding space of the diffusion model for inversion. Instead, we perform image inversion within the embedding space of a sentence transformer, facilitating a more clearer evaluation of the alignment between generated images and text prompts. Furthermore, our method does not require per-image optimization, which can be computationally expensive.

\noindent
\textbf{Curriculum learning.} Since we employ a novel curriculum learning regime to boost the performance of the studied models, we can also consider work on curriculum learning as related. The research community has extensively utilized this learning paradigm across a range of domains, including both computer vision \citep{Bengio-ICML-2009, Croitoru-arXiv-2022, Ionescu-CVPR-2016, Shi-ECCV-2016, Soviany-CVIU-2021, Chen-ICCV-2015, Sinha-NIPS-2020, Zhang-NeurIPS-2021} and natural language processing \citep{Croitoru-arXiv-2022,  Liu-IJCAI-2018, Platanios-NAACL-2019}. However, given the unique nature of each application, distinct data organization approaches have been developed to ensure optimal results. For example, in vision, the number of objects in the image is one criterion \citep{Soviany-CVIU-2021, Shi-ECCV-2016}, while, in natural language processing, both word frequency \citep{Liu-IJCAI-2018} and sequence length \citep{Kocmi-RANLP-2017, Tay-ACL-2019, Zhang-ISPASS-2021} are utilized. Other contributions tried to avoid estimating sample difficulty by implementing curriculum learning on the model itself \citep{Jarca-ECAI-2024,Karras-ICLR-2018, Sinha-NIPS-2020, Croitoru-arXiv-2022}, or by selecting the samples dynamically, based on the performance of the model \citep{Kumar-ANIPS-2010, Jiang-AAAI-2015}. 
Different from related approaches based on ordering  data samples according to their difficulty \citep{Bengio-ICML-2009, Soviany-CVIU-2021, Shi-ECCV-2016}, we propose to employ a novel approach to assess the difficulty level. More specifically, we utilize the mean cosine similarity between the prompt embedding produced by the model and the ground-truth embedding vector, measured at various stages of the standard training process.


\section{Method}
\label{method}

\noindent
\textbf{Overview.}
Our approach for the image-to-prompt-embedding generation task employs an image encoder to extract image representations. These representations are then converted into sentence embeddings using dense layers. To train the encoder, we propose a novel training pipeline that integrates multiple novel components. An overview of the proposed training pipeline is shown in Figure~\ref{fig_pipeline}. We train our model using two objectives. The main objective estimates the
cosine distance between the predicted and the target sentence embeddings. The second
objective aims to accomplish a classification task by predicting whether the words from
a predefined vocabulary are present in the original prompt, given the generated image. Our pipeline also employs a curriculum learning strategy, filtering out complex examples in the initial training iterations. As the training process continues, we gradually increase the complexity threshold, until we encompass all training samples. 



\noindent
\textbf{Main objective.} Our goal is to understand the generative process of text-to-image diffusion models. In particular, we choose one of the most representative diffusion models to date, namely Stable Diffusion \citep{rombach-CVPR-2022}. We argue that reversing the process, performing image-to-text generation, plays an important role in this regard. However, we do not focus on generating natural language descriptions of images. Instead, we concentrate on training models to predict embedding vectors as similar as possible to the actual embeddings obtained by applying a sentence transformer \citep{Reimers-EMNLP-2019} on the original prompts. This task holds comparable significance to predicting the original text, as recent studies \citep{morris-arxiv-2023} have demonstrated that embeddings from state-of-the-art text encoders can be almost entirely converted back into the original text. Furthermore,  
this indirect task 
avoids the use of the problematic text evaluation measures \citep{Callison-Burch-EACL-2006} commonly employed in image captioning, such as BLEU \citep{Papineni-ACL-2002}. Therefore, we map generated images to vectors that reside in the embedding space of a sentence transformer \citep{Reimers-EMNLP-2019}, which correspond to the actual prompts. \citet{Chen-ACL-2023} found that BERT-style encoders are better than CLIP-style encoders, when it comes to encoding text. We thus consider that the sentence BERT encoder generates more representative target embeddings for our task. Formally, let $(x_i, y_i)_{i=1}^{n}$ denote a set of $n$ training image and prompt pairs. Let $f_\theta$ and $s$ be a prompt generation model and a sentence transformer, respectively. Then, the main objective is to minimize the following loss:
\begin{equation}
\label{eq_task}
    \mathcal{L}_1 = \frac{1}{n} \sum_{i=1}^{n}{\left (1 - \frac{\langle f_\theta(x_i), s(y_i) \rangle}{\Vert f_\theta(x_i)\Vert \cdot \Vert s(y_i) \Vert}\right)},
\end{equation}
where $\theta$ represents the learnable parameters of the model.

\noindent
\textbf{Multi-label vocabulary classification.} Although the objective defined in Eq.~(\ref{eq_task}) maximizes the cosine similarity between the output and the target embedding, the training process does not directly exploit the prompt content in this setting. Therefore, we introduce an additional classification task, in which the model has to analyze the generated input image and determine if each word in a predefined vocabulary is present or not in the original prompt used to generate the image. We select the most frequent adjectives, nouns and verbs as elements of the vocabulary.
Formally, we create a vocabulary $V=\{t_1, \dots, t_m\}$ and, for each training prompt $y_i$, we build a binary vector $l_i = \left( l_{i1}, \dots, l_{im}\right)$ as follows:
\begin{equation}
\label{eq_multilabel}
    l_{ij} = \left\{
    \begin{array}{ll}
     1, \textnormal{if}\; t_j \in y_i \\
     0, \textnormal{otherwise}
\end{array} , \forall j \in \left\{1, ..., m\right\}.
    \right.
\end{equation}
To learn the frequent vocabulary words, we add an additional classification head, containing $m$ neurons, thus having one neuron per vocabulary item. Given a training set of $n$ image and prompt pairs $(x_i, y_i)_{i=1}^{n}$, we minimize the following objective with respect to $\hat{\theta}$:
\begin{equation}
\label{eq_multilabel_loss}
\begin{split}
    \mathcal{L}_2 &= \!\frac{1}{n\!\cdot\!m} \sum_{i=1}^n \sum_{j=1}^m l_{ij}\!\cdot\!\log{\!\left(1\!-\! \hat{l}_{ij}\right)}\! +\!(1\!-\!l_{ij})\!\cdot\! \log{\!\left(\hat{l}_{ij}\right)},
\end{split}
\end{equation}
where $\hat{l}_i =f_{\hat{\theta}}(x_i)$ is the vector of probabilities for all vocabulary items, and $\hat{\theta}$ consists of the parameters of the new classification head, as well as the trainable parameters $\theta$ of the underlying model.
We combine the losses from Eq.~(\ref{eq_task}) and Eq.~(\ref{eq_multilabel_loss}) into a single term:
\begin{equation}
\label{eq_final_loss}
    \mathcal{L} = \mathcal{L}_1 + \lambda \cdot \mathcal{L}_2,
\end{equation}
where $\lambda$ is a hyperparameter that controls the importance of the classification task with respect to the prompt estimation task. 

\begin{figure}[t]
\begin{center}
\centerline{\includegraphics[width=\linewidth]{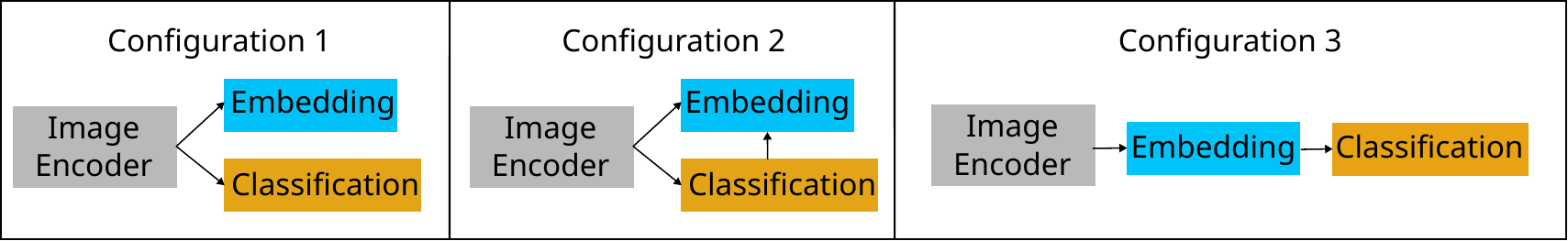}}
\vspace{-0.2cm}
\caption{Configurations for the classification and embedding prediction heads. In the first configuration, the heads are separate, being fed with the same features. In the second configuration, the output of the classification head is concatenated with the image encoding to create the final intake for the embedding prediction head. In the third configuration, the classification is carried out using the predicted embedding as input.}
\label{fig_heads}
\end{center}
\end{figure}

An important aspect for the classification task is the position of the new head relative to the head responsible for generating the prompt embedding, where each head is a single fully connected layer. We study three possible configurations, as illustrated in Figure~\ref{fig_heads}. In the first configuration, the vocabulary prediction head is added at the same level as the prompt embedding head, as an independent head, which can only influence the preceding representations of the model. In this configuration, the classification head can indirectly influence the prompt generation head. In the second configuration, the output of the classification head is passed as input to the embedding prediction head. This configuration aims to improve the prompt embedding by allowing the model to see which words are likely to be present in the prompt. In the third configuration, the model is constrained to produce an embedding from which the active words can be recovered. This configuration can better align the generated prompt with the actual prompt, when words from the vocabulary are present in  the original prompt. In the experiments, we empirically compare these three configurations.



\noindent
\textbf{Curriculum learning.} If there is a considerable amount of examples with noisy labels in the training data, the weights of our model could converge to suboptimal values during training. 
Since the input images are generated by a neural model, \ie~Stable Diffusion, we naturally expect to have images that do not fully represent the text prompts. Therefore, to make the training of the prompt generation models robust to noisy labels, we propose to employ a curriculum learning technique \citep{Bengio-ICML-2009}. Curriculum learning is a method to train neural models inspired by how humans learn \citep{Soviany-IJCV-2022}. It involves organizing the data samples from easy to hard, thus training models on gradually more difficult samples. In our case, we propose to
train the neural networks on samples with progressively higher levels of labeling noise. In the beginning, when the weights are randomly initialized, feeding easier-to-predict (less noisy) examples, then gradually introducing harder ones, can help convergence and stabilize training. 

Given that our outputs are embeddings in a vector space, we harness the cosine similarity between the generated prompt and the ground-truth prompt during training. Thus, we propose a two-stage learning procedure. In the first phase, we train the network for a number of epochs using the conventional learning regime, storing the cosine similarity of each sample after every epoch. The only goal of the first (preliminary) training phase is to quantify the difficulty of learning each training sample. The difficulty score of a training example is computed as the mean of the cosine similarities for the respective example. 
Generalizing the observations of \citet{Swayamdipta-EMNLP-2020}, we conjecture that the resulting difficulty score is proportional to the amount of noise, essentially quantifying the level of misalignment between the input image $x_i$ and the corresponding text prompt $y_i$. To support our conjecture, we show examples of easy, medium and hard images 
in Figure \ref{fig_cl_examples}.
For the second training phase, we reset the training environment and split the training data into three chunks, such that each chunk represents a different difficulty level: easy, medium and hard. Finally, we train the model again until convergence in the same number of steps as before, by gradually introducing each data chunk, starting with the easiest one and progressing to the hardest one. The model still gets to see the whole data set, but it ends up spending less time learning noisy examples.

We consider two alternative data splitting heuristics.
The first one is to divide the data set into three equally-sized chunks, inspired by \citet{Ionescu-CVPR-2016}. The second splitting heuristic involves setting two threshold values for the cosine similarity and splitting the data set according to these thresholds. As far as the latter heuristic is concerned, we propose a strategy for optimally choosing the threshold values. The first threshold should be set such that the first data chunk, \ie~the easy-to-predict samples, contains a significant portion of the data set. The second threshold should be sufficiently low to only include a small subset of the data set, \ie~the really difficult examples.


\noindent
\textbf{Underlying models.} To demonstrate the generalization power of our training framework across multiple pre-trained neural networks, we employ the proposed framework on four distinct deep architectures as image encoders: ViT \citep{Dosovitskiy-ICLR-2020} (Vision Transformer), CLIP \citep{radford-pmlr-2021} (Contrastive Language-Image Pre-training), Swin Transformer \citep{Ze-ICCV-2021}, and the encoder of the U-Net used in Stable Diffusion \citep{rombach-CVPR-2022}. The former three networks represent black-box models with no access to the weights of Stable Diffusion, while U-Net comes with the weights from Stable Diffusion, being a white-box approach.
%

%


The U-Net model operates in the latent space where the diffusion process takes place, and is based on a three-stage approach. First, we employ the perceptual compression auto-encoder used in Stable Diffusion to encode the images in the latent space. Next, we use an image captioning model \citep{Li-ICML-2022} to generate natural language descriptions. Lastly, based on the latent vectors and the captions, we train the encoder part of the U-Net architecture on our task, using the objective defined in Eq.~(\ref{eq_final_loss}). We use the captions similar to how the text is used for text-to-image generation, namely we integrate them in the cross-attention blocks of the U-Net.


\noindent
\textbf{Meta-regression.}
To further improve the performance of the 
ensemble, we introduce a domain-adaptive kernel method that allows the models to exploit information about the target (validation and test) domain without using the ground-truth embeddings. For a given similarity function $k$, we compute the kernel matrix $K$ containing similarities between pairs of samples from both source (training) and target (validation and test) domains. Thus, we have that:
\begin{equation}
\label{eq_kernel_matrix}
    K_{ij} = k(z_i, z_j), \forall i,j \in \left\{1, \dots, n+p \right\},
\end{equation}
where $Z = \left\{z_1, \dots, z_{n+p}\right\}$ denotes the union of the training set $X$ with an unlabeled data set $\bar{X}$ from the target domain, \ie~$Z = X \cup \bar{X} = \left\{x_1, \dots, x_{n}, \bar{x}_1 \dots \bar{x}_{p}\right\}$. Note that $n$ and $p$ indicate the number of examples in $X$ and $\bar{X}$, respectively. For simplicity, we use the linear kernel as the function $k$. Next, we normalize the matrix $K$ and utilize the radial basis function (RBF) to create a fresh set of second-order features for every sample, $\{z_i\}_{i=1}^{n+p}$. These features reflect the similarities between each sample and all other samples in the data set, including those present in the set representing the target domain. Formally, the kernel normalization can be performed via:
\begin{equation}
\label{eq_kernel_normalization}
\hat{K}_{ij} = \frac{K_{ij}}{\sqrt{K_{ii}\cdot K_{jj}}},
\end{equation}
while the RBF transformation is given by:
\begin{equation}
\label{eq_kernel_matrix_exp}
    {K_{\mbox{\scriptsize{DA}}}} = \exp{\left(-\gamma\left({1-\hat{K}_{ij}}\right)\right)}, 
\end{equation}
where $i,j \in \left\{1, \dots, n+p \right\}$.
To predict the prompt embeddings, we reapply the linear kernel on the revised features contained by $K_{\mbox{\scriptsize{DA}}}$ and train a dual regression model based on $L_2$ regularization. The final output of the regression model constitutes the improved prompt embedding of our framework. This approach offers a significant benefit, namely that the model has the opportunity to gather crucial information about the target domain in an unsupervised manner, by harnessing the similarities between source and target domain samples.
However, kernel methods are inefficient on large-scale data sets, due to their quadratic dependence on the number of training samples, which can lead to unreasonably high memory requirements. This statement directly applies to our work, as trying to compute a kernel matrix on our training set from DiffusionDB (even after filtering out near duplicates) produces an out-of-memory error (on a machine with 256 GB of RAM). To solve this issue, we employ the k-means algorithm to extract a set of centroids, denoted as $C=\left\{c_1, \dots c_r \right\}$, from our training set. Then, we substitute the original training set $X$ with this new set $C$ in the aforementioned method, which is significantly more efficient.

\section{Experiments}
\label{sec_experiments}

\subsection{Data set} 
DiffusionDB \citep{wang-arXiv-2022} contains 14 million images generated from 1.8 million prompts by Stable Diffusion \citep{rombach-CVPR-2022}. However, we observed that many prompts are near duplicates, and keeping all of them could make our task very easy (models could cheat by memorizing samples). 
To clean up the prompts, we introduce a set of successive filtering steps: $(i)$ eliminating any leading or trailing whitespaces, $(ii)$ removing examples having less than one word or containing \textit{Null} or \textit{NaN} characters, $(iii)$ removing examples containing non-English characters, and $(iv)$ discarding duplicate prompts having at least 50 identical characters at the beginning or the end. After applying the proposed filtering steps, we obtain a clean data set of approximately 730,000 image-prompt pairs. We divide the pairs into 670,000 for training, 30,000 for validation and 30,000 for testing.

\subsection{Hyperparameter tuning} 

We establish the hyperparameters during preliminary experiments on the validation set. Due to the large scale of the training set, we train all models for three epochs on an Nvidia GeForce GTX 3090 GPU with 24 GB of VRAM. The image resolution is either 224$\times$224 or 256$\times$256, depending on the model, while the mini-batch size is set to 64. The models are optimized with Adam. We set the learning rate to $10^{-4}$ for ViT/Swin models, and $5\cdot 10^{-6}$ for CLIP/BLIP-based  models. 
With the introduction of the classification head, there are two extra hyperparameters. These are the weight $\lambda$ of the additional classification loss and the size of the vocabulary $m$. We set $\lambda=0.1$ and $m=1000$. For DAKL, we set the number of k-means clusters to 10K and $\gamma=0.001$. To foster future research and allow others to fully reproduce our results, we release our code as open source\footnote{\url{https://github.com/CroitoruAlin/Reverse-Stable-Diffusion}}. 

\begin{table}[t]
\centering 
\setlength\tabcolsep{4.0pt}
\small{
\begin{tabular}{|c | c | c | c |} 
\hline
 {Image encoder} &  MLC
 & CL 
 & Cosine similarity \\
 \hline
 \hline
 CLIP-Huge & - & - 
 & 0.6725 \\
 CLIP-Huge (ours) & \checkmark & - 
 &0.6739 \\
 CLIP-Huge (ours) & \checkmark & \checkmark 
 & \textbf{0.6750} \\
 \hline
 U-Net$_{\mbox{\scriptsize{enc}}}$ & - & - 
 &0.6130 \\
 U-Net$_{\mbox{\scriptsize{enc}}}$ (ours) & \checkmark & - 
 & 0.6455 \\
U-Net$_{\mbox{\scriptsize{enc}}}$ (ours) & \checkmark & \checkmark 
& \textbf{0.6497} \\
\hline
 Swin-L & - & - 
 & 0.6624 \\
 Swin-L (ours) & \checkmark & - 
 & \textbf{0.6671} \\
 Swin-L (ours) & \checkmark & \checkmark 
 & \textbf{0.6671} \\
\hline
 ViT & - & - 
 & 0.6526 \\
 ViT (ours) & \checkmark & - 
 & 0.6539 \\
 ViT (ours) & \checkmark & \checkmark 
 & \textbf{0.6550} \\
 \hline

\end{tabular}
}
\vspace{-0.2cm}
\caption{Average cosine similarity scores between predicted and ground-truth prompt embeddings, employing different neural architectures, while gradually adding our novel components, namely the vocabulary multi-label classification (MLC) head and the curriculum learning (CL) procedure, to illustrate their benefits. The best score for each architecture is highlighted in bold. 
}
\label{tab_ablation} 
\end{table}

\begin{table}[t]
\centering 
\small{
\begin{tabular}{|c | c | c | c |c|} 
\hline
 & \multirow{ 2}{*}{Baseline} & \multicolumn{3}{c|}{Head configuration} \\
 \cline{3-5}
 & & 1 & 2 & 3\\
 \hline
 \hline
 Cosine& \multirow{ 2}{*}{0.6624}  &\multirow{ 2}{*}{0.6620} & \multirow{ 2}{*}{0.6632} & \multirow{ 2}{*}{\textbf{0.6671}} \\
 similarity & & & & \\
\hline
\end{tabular}
}
\vspace{-0.2cm}
\caption{Ablation study on the head configurations used for the classification task. The options illustrated in Figure~\ref{fig_heads} are tested on the Swin-L backbone.}
\label{tab_ablation_heads} 
\end{table}

\subsection{Prompt embedding prediction results}

In Table~\ref{tab_ablation}, we report the average cosine similarity scores between the prompt embeddings predicted by various neural models and the ground-truth prompt embeddings in our test set. We compare vanilla models with versions that are enhanced by our novel training pipeline. We introduce our novel components one by one, to illustrate the benefit of each added component.


\begin{table}[t!]
\centering 
\small{
\begin{tabular}{|c | c|  c | c |} 
\hline
 Model & MLC 
 & CL 
 & RefCLIPScore \\
 \hline
 \hline
BLIP& - & - & 30.93 \\
BLIP (ours)& \checkmark & - & 31.77\\
BLIP (ours)& \checkmark & \checkmark & \textbf{31.88}$^\ddagger$ \\
\hline
BLIP-2& - & - & 31.22 \\
BLIP-2 (ours)& \checkmark & - & 31.96\\
BLIP-2 (ours)& \checkmark & \checkmark & \textbf{32.14}$^\ddagger$ \\
\hline
\end{tabular}
}
\vspace{-0.2cm}
\caption{Comparison between fine-tuned vanilla BLIP and BLIP-2 models and the enhanced versions of these models. Our enhanced versions of BLIP and BLIP-2 integrate our multi-label classification (MLC) head and curriculum learning (CL) strategy. The best score for each architecture is highlighted in bold. Results marked with $\ddagger$ indicate that the corresponding models are significantly better than the reference vanilla models, according to a paired sample t-test, with a p-value lower than $\mathbf{0.001}$.}
\label{tab_image_caption} 
\end{table}

\begin{table}[t]

\vspace{-0.cm}
\centering 
\small{
\begin{tabular}{|c | c | c | } 
\hline
 {Method} &  {RefCLIPScore} &  {CLIPScore} \\
 \hline
 \hline
 PH2P \citep{Mahajan-CVPR-2024} & 25.13 & 0.7214 \\
\hline
 PEZ \citep{Wen-NeurIPS-2023} & 26.14 & 0.7458\\
\hline
 BLIP-2 (ours) &  \textbf{32.19} & \textbf{0.7468} \\
\hline
\end{tabular}
}
\vspace{-0.2cm}
\caption{Comparison between two state-of-the-art methods (PH2P and PEZ) versus our enhanced BLIP-2, on the prompt generation task (in terms of RefCLIPScore) and image regeneration task (in terms of CLIPScore), respectively. The best score for each measure is highlighted in bold.}
\label{tab_comparison_related_work} 
\vspace{-0.2cm}
\end{table}

\begin{figure*}[t!]
\begin{center}
\centerline{\includegraphics[width=1.0\linewidth]{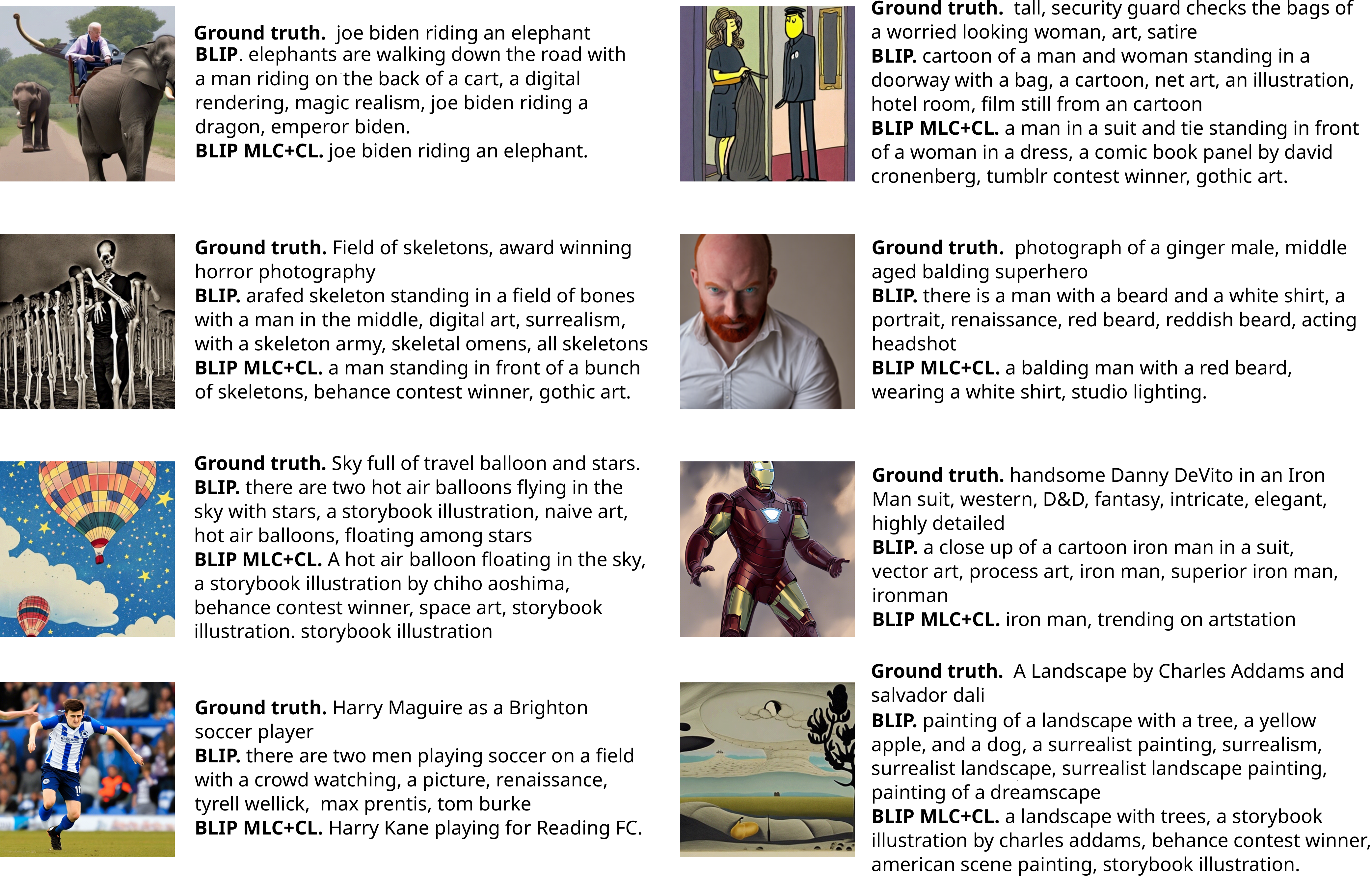}}
\caption{Examples of captions for generated images. We compare the prompts returned by a fine-tuned vanilla BLIP with those of 
an enhanced version of BLIP based on multi-label classification (MLC) and curriculum learning (CL). Best viewed in color.}
\label{fig_prompt_caption}
\vspace{-0.5cm}
\end{center}
\end{figure*}

As shown in Table~\ref{tab_ablation}, our findings indicate that the classification head leads to consistent performance gains. For three out of four models, our curriculum learning strategy brings further performance improvements. 
We thus conclude that all our novel contributions boost performance. Notably, our highest gains are observed for the U-Net$_{\mbox{\scriptsize{enc}}}$ model, as the performance grows from $0.6130$ to $0.6497$.

\noindent
\textbf{Ablation study.}
In Table~\ref{tab_ablation_heads}, we present the results obtained by each head configuration illustrated in Figure~\ref{fig_heads}. We observe that the third option gives the best results, showing that it is useful to update the embedding head based on the classification error. This result strengthens the claim that our additional classification task is useful in estimating the sentence embeddings.

\subsection{Prompt generation results}

\noindent
\textbf{Quantitative results.}
Although we study the task of prompt embedding prediction, which is a well-posed reverse engineering task, predicting the actual prompt is also possible, but we consider this latter task ill-posed. 
To this end, we further employ our framework based on curriculum learning (CL) and an extra multi-label classification (MLC) head to fine-tune the BLIP and BLIP-2 models on the image captioning task. We next perform a quantitative analysis on the entire test set to compare our approaches with the vanilla BLIP and BLIP-2 models. As evaluation metric, we use the reference-augmented version of the recently proposed CLIPScore \citep{Hessel-EMNLP-2021}, namely the RefCLIPScore. The corresponding results are presented in Table \ref{tab_image_caption}. 

The BLIP model is a state-of-the-art captioning model that obtains a RefCLIPScore of 30.93. 
The proposed version of BLIP based on multi-label classification and curriculum learning yields superior results, increasing the RefCLIPScore of BLIP from 30.93 to 31.88. We observe a similar trend for BLIP-2, where the RefCLIPScore increases from 31.22 to 32.14. Additionally, we perform paired sample t-tests to compare the fine-tuned vanilla BLIP and BLIP-2 models against our enhanced versions of these models. The statistical tests indicate that the improvements brought by our framework are significant, with p-values below $0.001$.
These results show that our framework can significantly improve image captioning results, thus extending its applicability from prompt embedding generation to generated image captioning.

\noindent
\textbf{Qualitative evaluation of generated captions.}
We qualitatively compare the captions returned by our enhanced BLIP and those of the fine-tuned vanilla BLIP. We display a set of representative samples in Figure~\ref{fig_prompt_caption}. 
For the less complex images, a matching prompt is usually found by the vanilla BLIP. The predicted prompts for the harder examples are still representative and depict certain nuances of the text, but they often fail to precisely describe all aspects of the input images. 
Our enhanced version of BLIP (based on MLC+CL) produces improved prompts in a number of cases, \eg~the first image on the first column, the second image on the second column, or the last image on the second column. Although the prompts of our best model are representative, they are still far from the ground-truth prompts, suggesting that the generated image captioning task is indeed ill-posed. A representative ill-posed case is the third image on the second column, depicting Iron Man, where it is impossible to predict the prompt, as Danny DeVito is hidden by the Iron Man suit. Indeed, there is no visual clue to indicate the presence of Danny DeVito in the source prompt.

\noindent
\textbf{Related work comparison.} We consider PH2P~\citep{Mahajan-CVPR-2024} and PEZ~\citep{Wen-NeurIPS-2023} as the most closely related approaches to our work. Both methods focus on generating the most likely prompt for a given image that could have been used to create the respective image. Therefore, we compare our prompt generation results against these two related methods, PH2P~\citep{Mahajan-CVPR-2024} and PEZ~\citep{Wen-NeurIPS-2023}, in Table~\ref{tab_comparison_related_work}. For this comparison, we employ two metrics. On the one hand, we use the RefCLIPScore for prompt quality evaluation. On the other hand, we report the CLIP score calculated in the image space, comparing the original and regenerated images for each method, independently. We perform the comparison on 100 examples because PH2P and PEZ are per-image optimization methods, being implicitly slow on generating the prompts. By analyzing the results, we observe that the prompts generated by our BLIP-2 model not only align closer to the original prompt, but also effectively produce images that closely resemble the target images.
\begin{figure*}[!t]
\begin{center}
\centerline{\includegraphics[width=1\linewidth]{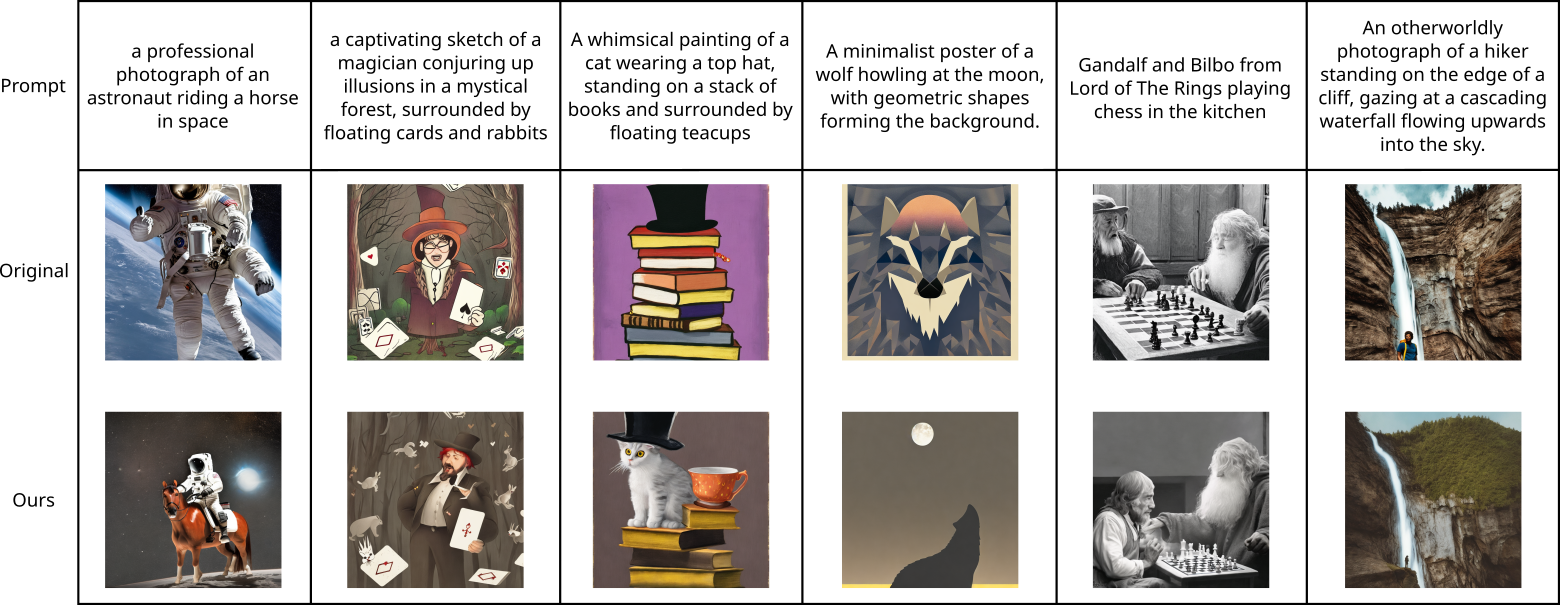}} 
\vspace{-0.2cm}
\caption{Samples generated by original and modified Stable Diffusion models. The images on the middle row are synthesized by the original U-Net. The images on the bottom row are generated by replacing (from the second diffusion step onward) the original U-Net encoder with our U-Net encoder employed for prompt embedding prediction. 
Notable differences include the presence of a horse and rabbits in the bottom images, while they are absent in the top ones (first and second column). Our model also corrects errors like the orientation of a person (last column) and the height of characters like Bilbo, recognizing him as a hobbit and adjusting his height accordingly. Best viewed in color.
} 
\label{fig_generation_results}
\vspace{-0.3cm}
\end{center}
\end{figure*}

\begin{table*}[t]

\vspace{-0.cm}
\centering 
\small{
\begin{tabular}{|c | c | c | c | c| c|} 
\hline
 {Choice} &  {Person \#1} &  {Person \#2} & {Person \#3} & Agreements & Average\\
 \hline
 \hline
 Ours & 52 & 36 & 31 & 21 & 39.6 \\
 \hline
 Original& 30 & 10 & 5 & 2 & 15.0\\
\hline
 Undecided & 18 & 54 & 64 & 14 & 45.4\\
\hline
\end{tabular}
}
\vspace{-0.2cm}
\caption{Results of the subjective human evaluation study on image-text alignment completed by three volunteers. Each annotator was shown 100 image pairs and asked to choose the image that best aligns with the given prompt. For each displayed pair, the left or right image locations were randomly chosen to avoid cheating. Each person had a third option if they were unsure, denoted as \emph{undecided}.}
\label{tab_img_gen} 
\vspace{-0.2cm}
\end{table*}


\subsection{Application to image generation}
\label{sec_img_gen}
We next demonstrate the promising capabilities of our framework. We specifically illustrate the ability to synthesize images that are better aligned with the prompts. This is achieved by simply replacing the U-Net in the Stable Diffusion pipeline with the one used in our prompt prediction task. Our initial attempt resulted in spurious invalid images, \ie~very similar to noise. Therefore, for the first iteration only, we use the original U-Net, then switch to our U-Net for the remaining steps of the diffusion process. This approach produces more faithful visual representations of the prompts. Figure~\ref{fig_generation_results} depicts a few qualitative results, where our model exhibits better alignment between
the generated image and the text prompt. We observe that the images generated by
our model illustrate more details that are omitted in the first instance by the original U-Net.

To ensure a fair comparison of the original and modified Stable Diffusion models, we conducted an experiment based on human feedback. The experiment involved three individuals who evaluated 100 pairs of randomly sampled images, from an image-to-text alignment perspective. Each participant was given the same set of 100 image pairs. Each pair is formed of an image generated by our U-Net, and another one, generated by the original model. Both images in a pair are generated for the same prompt. The generative process ran $50$ denoising steps with the DDIM sampler \citep{song-ICLR-2021b}. The participants had to choose the image that is more representative for the given prompt, having three options: first image, second image, and undecided. A pair labeled as \emph{undecided} denotes the case when a participant is not able to choose a better image. Moreover, the location of the images (left or right) is randomly assigned within each pair to prevent disclosing the source model that generated each image, thus avoiding any form of cheating. 

The detailed results of the above experiment are shown in Table~\ref{tab_img_gen}.
We report the number of votes awarded by each person to each of the two models, ours and the original, along with the number of undecided votes by each participant. Overall, the voting results show that, indeed, the usage of our U-Net in the generative process yields more text-representative images.
Furthermore, we also compute the number of times the three participants agree on the same response, and we save the results in the fifth column. The agreement results are also favorable to our U-Net. 


\begin{figure*}[t!]
\begin{center}
\centerline{\includegraphics[width=0.9\linewidth]{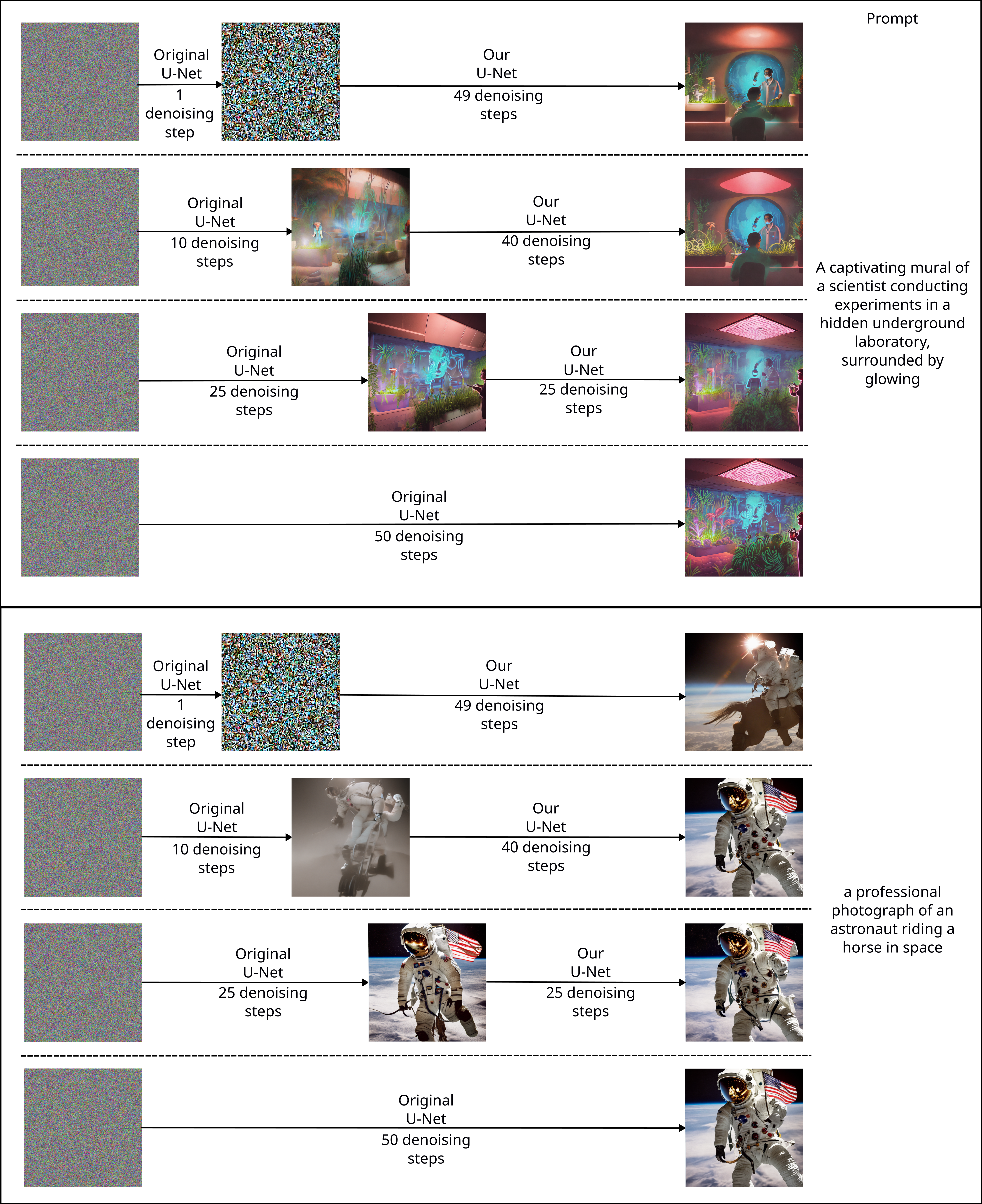}}
\vspace{-0.2cm}
\caption{Examples that demonstrate the impact of replacing the original U-Net in Stable Diffusion with our U-Net on text prompt generation, at various steps of the denoising diffusion process. Best viewed in color.}
\label{fig_denoising_steps}
\end{center}
\end{figure*}

In addition to the human evaluation experiment, we perform a quantitative analysis to assess the performance of our fine-tuned U-Net in terms of its alignment with the conditional prompts. In Table~\ref{tab_quantitative_experiments}, we report the cosine similarities between the sentence embeddings predicted by our fine-tuned U-Net and the ground-truth embeddings on DrawBench \citep{saharia-arXiv-2022}, as well as on a subset of 2500 test examples from DiffusionDB. 
We observe that our U-Net outperforms the original model across both data sets. Notably, on DrawBench, the gap in favor of our version of U-Net is significant.


\begin{table}[!t]
\centering 
\small{
\begin{tabular}{|c | c | c |} 
\hline
Model & DrawBench  & DiffusionDB \\
 \hline
 \hline
Original U-Net & 0.4594	&0.6417 \\
\hline
Our U-Net & 0.5067	& 0.6443\\
\hline
\end{tabular}
}
\caption{Comparison between our U-Net and the original U-Net on DrawBench and DiffusionDB. After generating the images, we compute the cosine similarity between the embedding predicted by the model and the ground-truth embedding. For DiffusionDB, we run the experiment on 2500 samples that are randomly selected from the test prompts.}
\label{tab_quantitative_experiments}  
\end{table}

\noindent
\textbf{When to replace U-Net in Stable Diffusion?}
We hereby present a qualitative analysis of the impact of the denoising step at which we substitute the original U-Net with ours. To illustrate how this change affects the final samples, we showcase two prompt examples and several images in Figure~\ref{fig_denoising_steps}. We vary the number of denoising steps performed by our U-Net. Specifically, we use it to perform the last $49$, $40$, and $25$ denoising steps, respectively. The illustrated examples indicate that the optimal text-to-image alignment is achieved for both prompt examples, when only a single denoising step is performed with the original U-Net. However, when we make the switch in the later stages of the denoising diffusion process, the impact on the final image becomes less meaningful. In the first example, when we introduce the model after the first $10$ steps, the output is still aligned with the text, but for the second example, the horse is removed when performing $10$ steps with the original model. Overall, we conclude that the first part of the denoising process has the highest impact on the content of the final image. When we switch the model in the second half of the denoising process, the results are very similar to the case when we use only the original U-Net. Based on these observations, we decided to replace the original U-Net with our own right after the first sampling step in our application to image generation discussed above.

\begin{figure}[t!]
\begin{center}
\centerline{\includegraphics[width=0.7\linewidth]{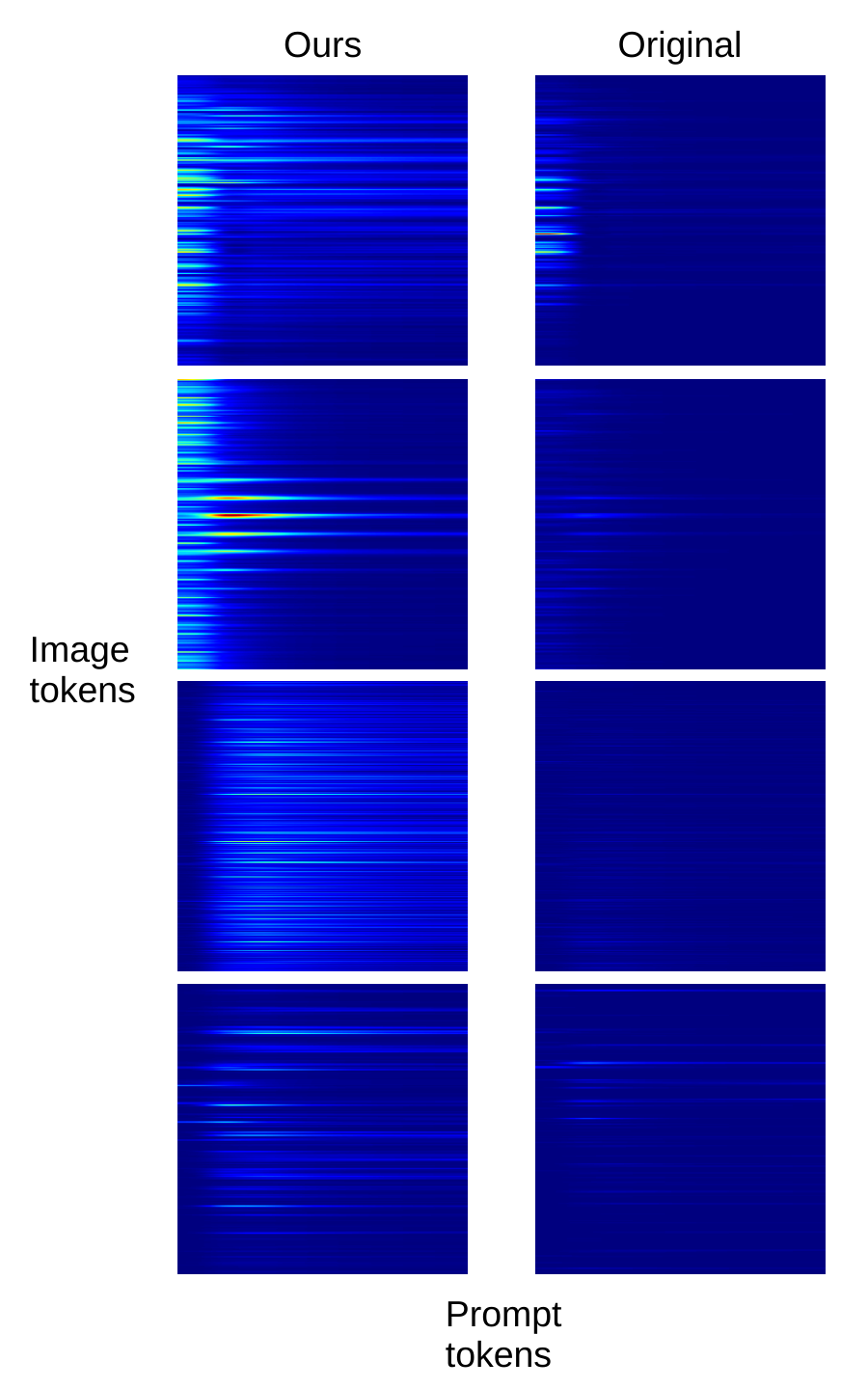}}
\vspace{-0.2cm}
\caption{Comparison of cross-attention maps between the original and our fine-tuned U-Net. Best viewed in color.}
\label{fig_att_maps}
\end{center}
\end{figure}

\noindent
\textbf{Cross-attention maps of original and fine-tuned U-Nets.}
In Stable Diffusion, the conditional prompts are integrated in the pipeline via cross-attention layers. We analyze the cross-attention maps of these layers and observe that our fine-tuned U-Net covers more content words compared to the original model. We provide some illustrative examples of cross-attention maps in Figure~\ref{fig_att_maps}. For improved visualization, we only keep the tokens corresponding to content words and adjust the images to a 1:1 aspect ratio.

\begin{figure*}[t!]
\begin{center}
\centerline{\includegraphics[width=0.68\linewidth]{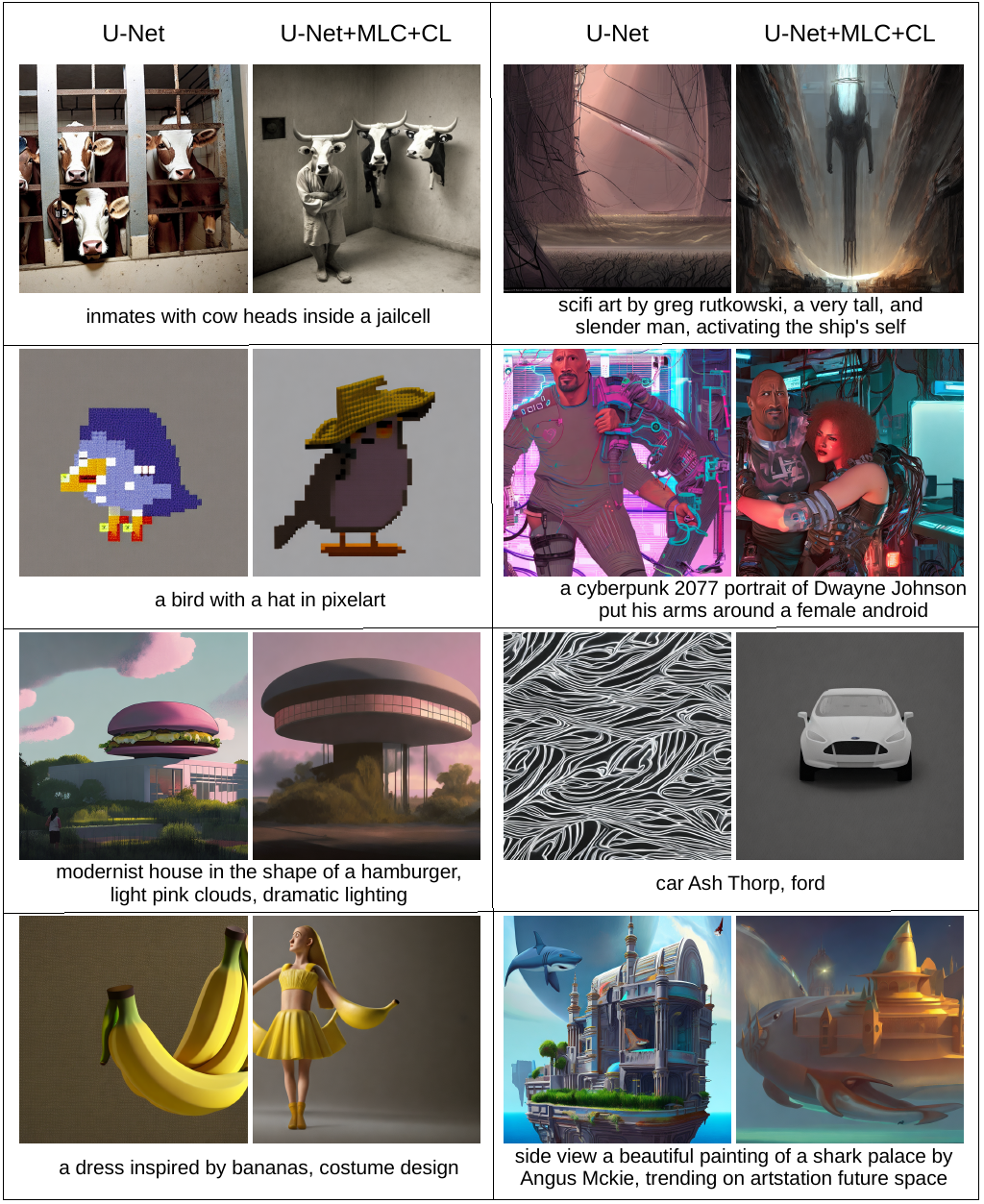}}
\vspace{-0.25cm}
\caption{Images generated by the fine-tuned U-Net, before and after adding our novel components, namely the multi-label classification (MLC) head and the curriculum learning (CL) strategy. Best viewed in color.}
\label{fig_unet_ba_mlc_cl}
\end{center}
\end{figure*}

\noindent
\textbf{Qualitative evaluation of the novel components added to U-Net.}
To study the effect of our novel components on the alignment of images generated by the fine-tune U-Net, we generate several images with the fine-tuned U-Net, before and after introducing our extra techniques, namely the multi-label classification head and the curriculum learning strategy. We present the images in Figure \ref{fig_unet_ba_mlc_cl}. The figure shows several cases where the enhanced version of U-Net produces images that are better aligned with the prompts. For example, in the right-hand side example from the first row, the vanilla fine-tuned U-Net does not generate the man. Similarly, in the left-hand side example from the second row, it does not generate the hat on the bird's head. The vanilla model also fails to generate the female android in the right-hand side example from the second row. In all these cases, our enhanced version of U-Net generates the objects referred in the prompts, confirming the benefits of adding our novel components. Note that, in Table \ref{tab_img_gen}, we show quantitative results demonstrating that the enhanced U-Net achieves higher cosine similarity scores. Hence, the qualitative results shown in Figure \ref{fig_unet_ba_mlc_cl} are consistent with the quantitative results.

In summary, the outcome of the above experiment is remarkable, given that our U-Net is trained on prompt prediction and not on image generation. It is clear that our model performs better when evaluated from the text-to-image alignment perspective.


\subsection{Qualitative analysis of the proposed difficulty score}

\begin{figure*}[!t]
\begin{center}
\centerline{\includegraphics[width=1.\linewidth]{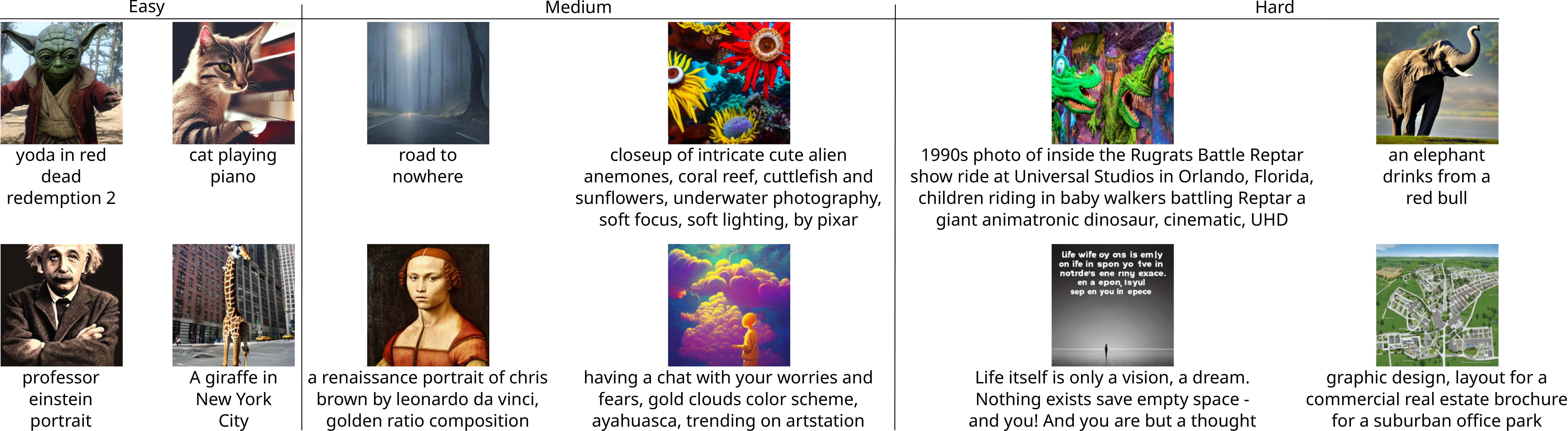}}
\vspace{-0.1cm}
\caption{Examples labeled as easy, medium and hard by our difficulty estimation procedure based on monitoring the cosine similarity of samples during conventional training. Best viewed in color.}
\label{fig_cl_examples}
\vspace{-0.4cm}
\end{center}
\end{figure*}

In Figure \ref{fig_cl_examples}, we illustrate some examples showing how our method categorizes image-prompt pairs based on their difficulty (easy, medium, or hard). The easy samples contain short straight-forward descriptions, which are very well aligned with the generated images. The medium examples involve descriptions that produce abstract images, or descriptions that require rich creativity. Finally, the hard samples consist of prompts that cannot be associated with a visual representation, \eg~quotes, or very complex prompts with multiple, especially unreal, elements. These examples indicate that our difficulty scores are easy to interpret visually, suggesting a reasonable organization of the easy, medium and hard example batches, which is correlated with the alignment between images and prompts.

\subsection{Measuring the effect of input noise}

\begin{figure*}[!t]
\begin{center}
\centerline{\includegraphics[width=1\linewidth]{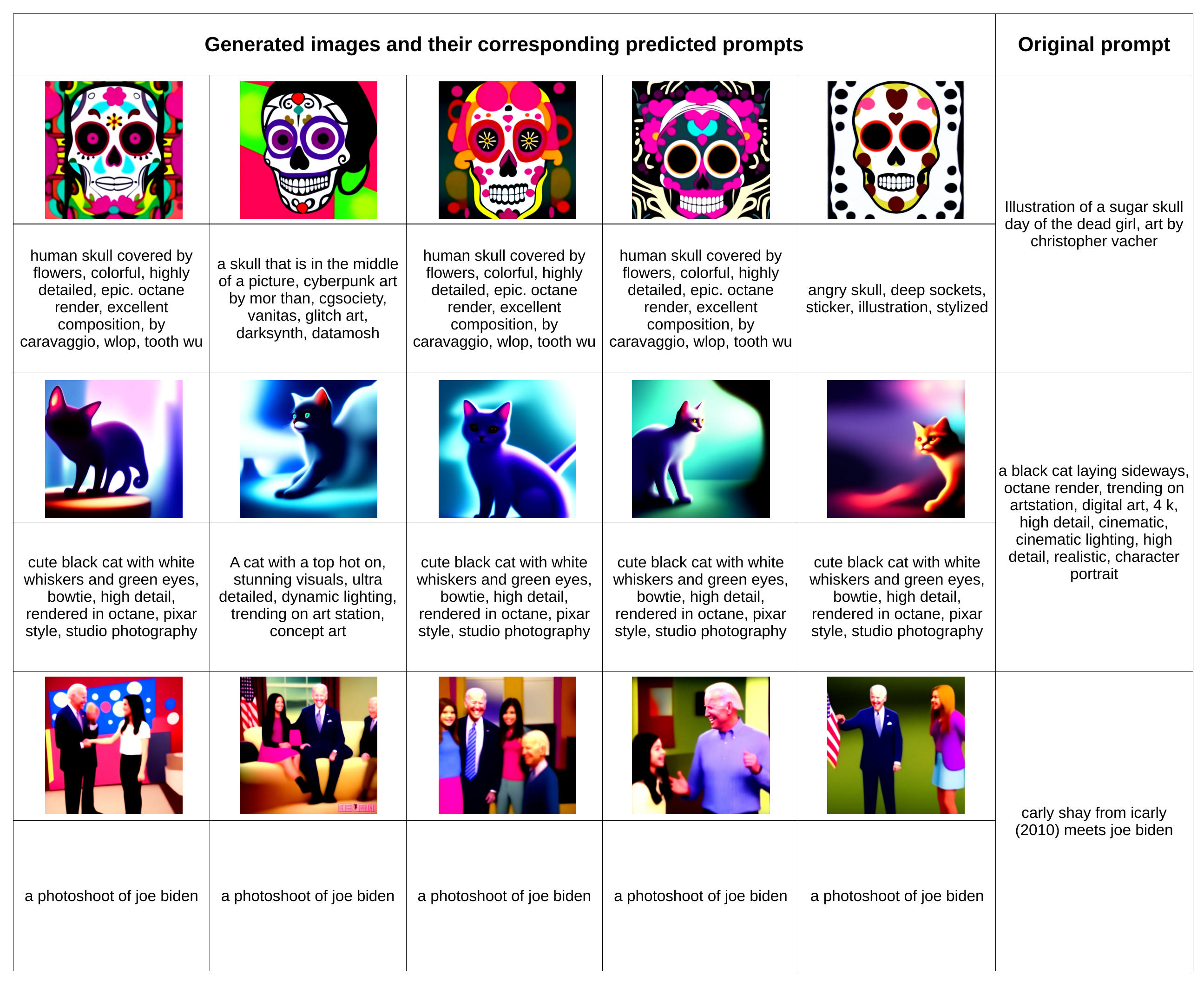}} 
\vspace{-0.2cm}
\caption{Generated images starting from a different input noise, as well as their predicted prompt by our enhanced BLIP-2. Best viewed in color.}
\label{fig_random_noise}
\vspace{-0.3cm}
\end{center}
\end{figure*}

We further investigate the effect of the input noise from which the backward diffusion process starts. In order to assess this effect, we carry out both quantitative and qualitative experiments.

\begin{table}[!t]
\centering 
\small{
\begin{tabular}{|c | c | c | c |} 
\hline
Backbone & Curriculum 
& Cosine\\
& learning method & similarity \\
 \hline
  \hline
\multirow{4}{*}{ViT} & No curriculum & 0.6526 \\
 \cline{2-3}
& LeRaC \citep{Croitoru-arXiv-2022} & 0.6525 \\
& CBS \citep{Sinha-NIPS-2020} & 0.6254 \\
\cline{2-3}
& Ours & \textbf{0.6544} \\
\hline
\end{tabular}
}
\caption{Comparison between our curriculum learning method and two state-of-the-art curriculum learning approaches, CBS \citep{Sinha-NIPS-2020} and LeRaC \citep{Croitoru-arXiv-2022}. The best score is highlighted in bold.}
\label{tab_curriculum_compare}  
\end{table}

For the quantitative assessment, we generate 10 images for each prompt from a set of 100 test prompts, using a different input noise as a starting point for each generated image. Next, we use our model to predict the BERT embedding for each generated image. For each prompt, we further compute the mean embedding vector for the 10 images. Then, we compute the standard deviation of the cosine similarities between each embedding and the mean embedding vector. We average the standard deviations over all prompts, resulting in an average standard deviation of $0.0175$. This indicates that the generated BERT embeddings exhibit marginal fluctuations when varying the input noise.

For the qualitative evaluation, we employ Stable Diffusion to generate five images per prompt, starting from a different input noise each time a new image is generated. Then, for each synthetic image, we predict the prompt using our enhanced BLIP-2 model. We present generated samples for three prompts in Figure \ref{fig_random_noise}. The illustrated examples confirm that the predicted prompts are consistent with each other, demonstrating that the output embeddings vary only by marginal amounts. In conclusion, we attest that our method is not significantly influenced by the input noise.

\begin{figure*}[!t]
\begin{center}
\centerline{\includegraphics[width=1.0\linewidth]{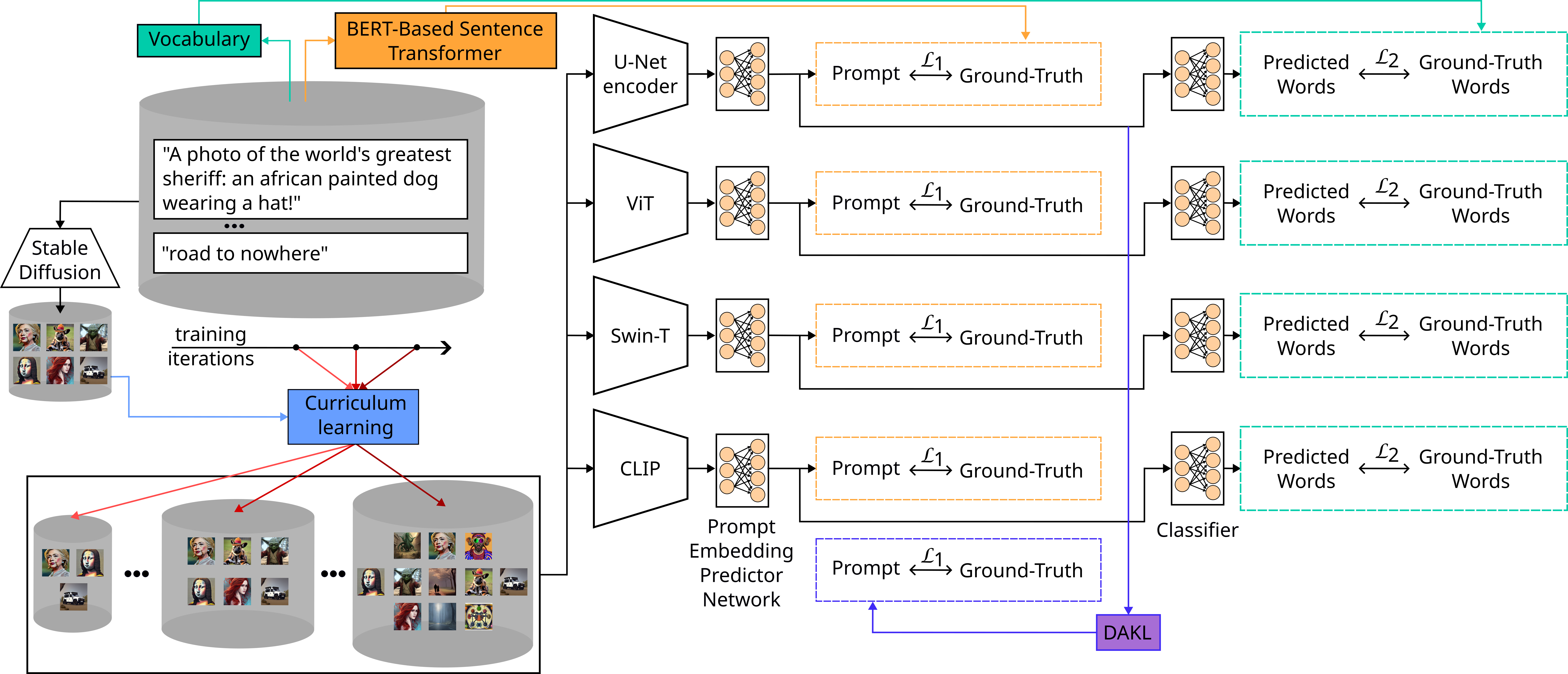}}
\vspace{-0.2cm}
\caption{Our prompt embedding estimation ensemble framework, which includes a classification task, a curriculum learning procedure and a domain-adaptive kernel learning (DAKL) method. As part of the curriculum learning procedure, we start with a small set of easy examples and gradually add more complex samples in the subsequent iterations. We transform the input prompts via a sentence transformer for the embedding estimation task, and we use a vocabulary of the most common words to create the target vectors for the classification task. Lastly, we apply the DAKL method on the median embedding of the four models to further improve the performance of the framework. Best viewed in color.}
\label{fig_pipeline_full}
\end{center}
\end{figure*}

\begin{table*}[t]

\vspace{-0.2cm}
\centering 
\setlength\tabcolsep{3.0pt}
\small{
\begin{tabular}{|c | c | c | c
| c |} 
\hline
 {Image encoder} &  Multi-label classification
 & Curriculum
  learning 
 & {DAKL}
 & Cosine similarity \\
 \hline
\hline
Ensemble & \checkmark & - & - 
 & 0.6879 \\
Ensemble & \checkmark & \checkmark & - 
& 0.6900 \\
Ensemble &\checkmark & \checkmark & \checkmark
&  \textbf{0.6917} \\
\hline
\end{tabular}
}
\caption{Average cosine similarity scores between predicted and ground-truth prompt embeddings, employing all the models from Table~\ref{tab_ablation}, while gradually adding our novel components, namely the vocabulary classification head and the curriculum learning procedure, to illustrate their benefits. We also report the results before and after integrating our DAKL method.}
\label{tab_results} 
\end{table*}

\subsection{Comparison with other curriculum learning methods}

To assess the performance of our proposed curriculum learning technique, we compare it with other curriculum learning methods from the recent literature. We choose two state-of-the-art methods, namely Curriculum by Smoothing (CBS) \citep{Sinha-NIPS-2020} and Learning Rate Curriculum (LeRaC) \citep{Croitoru-arXiv-2022}. Each of these methods have additional hyperparameters, which we tried to carefully tune via grid search. Unfortunately, we did not manage to make them surpass the performance of the vanilla training regime. As shown in Table~\ref{tab_curriculum_compare}, the results demonstrate the net superiority of our curriculum learning method in image-to-prompt generation with the ViT backbone. The success of our technique relies on harnessing the misalignment level between the image and the prompt in each training pair to create the curriculum schedule. In contrast, CBS and LeRaC do not take the misalignment into account. We believe this explains why our results are better.


\subsection{Model ensembles}



\noindent
\textbf{Basic ensemble.}
To boost the performance of our method, we have implemented an ensemble comprising four independently trained models (ViT, CLIP, Swin-T and U-Net), as depicted in Figure \ref{fig_pipeline_full}. The output of this ensemble is calculated as the average of the prompt embeddings predicted by each of these four models. 

Since our goal is to assess how well the text embedding can be recovered from generated images, we motivate our use of an ensemble of multiple models via the focus on minimizing the possibility of reporting low cosine similarity scores due to a poor model choice for the reverse task.


\noindent

\begin{table*}[t]
\centering 
\setlength\tabcolsep{3.0pt}
\small{
\begin{tabular}{| c |c | c | c | c | c |} 
\hline
\#Models & CLIP-Huge &  $\;$U-Net$_{\mbox{\scriptsize{enc}}}\;$ & $\;\;\;$Swin-L$\;\;\;$ & $\;\;\;\;\;$ViT$\;\;\;\;\;$ & Cosine similarity \\
 \hline
 \hline
\multirow{4}{*}{1}& \checkmark & - & - & - & {0.6750} \\
& - & \checkmark & - & - & {0.6497} \\
& - & - & \checkmark & - & {0.6671} \\
& - & - & - & \checkmark & {0.6550} \\
\hline
\multirow{6}{*}{2} & \checkmark & \checkmark & - & - & {0.6785} \\
& \checkmark & - & \checkmark & - & {0.6887} \\
& \checkmark & - & - & \checkmark & {0.6854} \\
& - & \checkmark & \checkmark & - & {0.6787} \\
& - & \checkmark & - & \checkmark & {0.6732} \\
& - & - & \checkmark & \checkmark & {0.6792} \\
\hline
\multirow{4}{*}{3} & \checkmark & \checkmark & \checkmark & - & {0.6900} \\
& \checkmark & \checkmark & - & \checkmark & {0.6901} \\
& \checkmark & - & \checkmark & \checkmark & {0.6901} \\
& - & \checkmark & \checkmark & \checkmark & {0.6820} \\
\hline
4 & \checkmark & \checkmark & \checkmark & \checkmark & \textbf{0.6917} \\
\hline
\end{tabular}
}
\vspace{-0.2cm}
\caption{Cosine similarity scores between predicted and ground-truth prompt embeddings, while employing different combinations of neural architectures. Individual models are compared with combinations of two, three and four models. The top score is highlighted in bold.}
\label{tab_ablation_models} 
\end{table*}

\noindent
\textbf{Results.}
Comparing the results in Table \ref{tab_ablation} and Table \ref{tab_results}, we observe a noticeable gap between the individual models and the ensemble. We thus believe the scores reported for the ensemble better reflect the misalignment of the original Stable Diffusion model. Our domain-adaptive kernel learning (DAKL) method further boosts the performance of the ensemble. Below, we report ablation results with combinations of two and three models (see Table \ref{tab_ablation_models}), which clearly indicate that all individual models play a role in the proposed ensemble.

\noindent
\textbf{Ablating the ensemble.} 
To better motivate the proposed combination of models, we conduct additional experiments with various ablated combinations of models. The corresponding results are shown in Table \ref{tab_ablation_models}. We observe that combining every two models leads to better results than using the individual counterparts. Further performance gains are obtained by combining every three models. Still, the top cosine similarity is reached when we combine all four models. In conclusion, our results clearly indicate that all individual models contribute to improving the proposed ensemble.

Furthermore, we observe that the gains saturate each time we increase the number of models in the ensemble. This suggests that adding even more models would generate marginal gains. Thus, we limit ourselves to using the proposed ensemble based on four models.

To provide a more comprehensive overview of the hyperparameters of our DAKL method, we present additional results by varying the parameter $\sigma$ of the RBF transformation. Additionally, we explore various choices for the number of centroids $r$ used by the k-means algorithm. We present the corresponding empirical results in Table~\ref{tab_ablation_K}. There are multiple hyperparameter combinations surpassing the baseline, but the best results are obtained for $r=10,000$ clusters and $\gamma=0.001$.

\begin{table}[t]
\centering 
\small{
\begin{tabular}{|c | c | c | c |} 
\hline
DAKL & $r$ & $\gamma$  & Cosine similarity \\
 \hline
  \hline
- &  - & - & 0.6900\\
 \hline
\checkmark & 1,000 & 0.001 & 0.6899 \\
\checkmark & 5,000 & 0.001 & 0.6905 \\
\checkmark & 10,000 & 0.001 &0.6917 \\
 \hline
\checkmark & 10,000 & 0.0001 & 0.6898\\
\checkmark & 10,000 & 0.001 &0.6917 \\
\checkmark & 10,000 & 0.01 & 0.6909 \\
\checkmark & 10,000 & 0.1 & 0.6878  \\
\hline
\end{tabular}
}
\vspace{-0.2cm}
\caption{Varying the number of k-means clusters $r$ and the parameter $\gamma$ of the RBF kernel used in DAKL.}
\label{tab_ablation_K}  
\end{table}

\begin{table}[t]
\centering 
\small{
\begin{tabular}{|c |c| c |} 
\hline
 {Image encoder} & Type & Cosine similarity \\
 \hline
 \hline
CLIP-Huge + k-NN & $\blacksquare$ & 0.6189 \\
BLIP & $\blacksquare$ & 0.5129 \\
\hline
 CLIP-Huge & $\blacksquare$ & 0.6725 \\
 Swin-L & $\blacksquare$ & 0.6624 \\
 ViT & $\blacksquare$ & 0.6526 \\
\hline
 U-Net$_{\mbox{\scriptsize{enc}}}$ & $\square$ & 0.6130 \\
\hline
\end{tabular}
}
\vspace{-0.2cm}
\caption{Comparison between several neural architectures, which are divided into two categories: black box ($\blacksquare$) and white box ($\square$). Black-box models do not have access to the weights of Stable Diffusion. In contrast, the white-box model starts the fine-tuning process from the weights of Stable Diffusion.}
\label{tab_extra_results} 
\end{table}

\subsection{Results with other tested models}

\noindent
\textbf{Additional baselines.} 
We previously reported results with three models having no knowledge about the internals of Stable Diffusion, treating the diffusion model as a black box. We also employed the U-Net encoder from Stable Diffusion, which comes with the pre-trained weights of the diffusion model. Hence, we consider the approach based on U-Net as a white-box method. As underlying models, we initially considered two more black-box architectures. The first one is a k-nearest neighbors (k-NN) algorithm, applied in a regression setting. Leveraging the power of a fine-tuned CLIP to match the image and text representations, the embedding of the image is compared to all the embeddings obtained from reference training prompts. Then, based on the distance to the closest neighbors, the output embedding (in the sentence transformer space) is regressed as a weighted mean. The second baseline is represented by a BLIP model \citep{Li-ICML-2022}, a recent approach with state-of-the-art results in image captioning, which is fine-tuned on our task. 

\noindent
\textbf{Results.} In Table~\ref{tab_extra_results}, we compare the previous four models 
with the two additional models. For a fair comparison, all models are trained with the vanilla training procedure. We emphasize that the three black-box models (CLIP-Huge, Swin-L, and ViT) chosen as underlying architectures for our novel training framework are the most competitive ones. Hence, increasing the performance levels of these models by employing our learning framework is more challenging. This is why our learning framework exhibits the highest performance boost for the U-Net encoder, which starts from a lower average cosine similarity compared with the top three black-box models.

Another interesting observation is that the white-box U-Net is not necessarily the best model. Indeed, the privilege of having access to the weights of the Stable Diffusion model seems to fade out in front of very deep architectures, such as Swin-L and CLIP-Huge, that benefit from large-scale pre-training.

\section{Discussion}

Our contribution is a just step towards understanding diffusion models, and we acknowledge that there are many things to be addressed in order to completely solve the understanding of diffusion models. Nevertheless, our study contributes to the better understanding and use of diffusion models in several ways, thus representing a stepping stone to many improvements in image generation. 

First, the generated captions can be manually inspected and compared with the original captions, to understand which concepts were integrated by Stable Diffusion in the generated image, and which concepts are left out. This can help the user to adjust the prompt in order to obtain images that are more representative. 

Second, our study elucidates if one can reverse the text-to-image diffusion process. To address this task, we developed a powerful ensemble based on several models. The cosine similarity values present in Tables \ref{tab_ablation} and \ref{tab_results} range between 0.65 and 0.7, indicating that the text-to-image diffusion process can be reversed, only to some degree. The results indicate that, in the original diffusion model, there is a misalignment between the original text prompts and the generated images. Hence, our study shows a way to quantify the misalignment gap.

Third, our framework can directly be used to enhance text-to-image generation results, as shown by the quantitative and qualitative results discussed in the Section \ref{sec_img_gen}.

\section{Conclusion}
\vspace{-0.1cm}

In this paper, we explored the task of recovering the embedding of a prompt used to generate an image, which represents the reverse process of the well-known text-to-image generation task. We proposed a joint training pipeline that integrates three novel components, and showed how they contribute to a higher performance. These are: a multi-label classification head for the most frequent words, a curriculum learning scheme for improved convergence, and a domain-adaptive kernel learning framework for meta-regression. In our experiments, we leveraged the large number of image-prompt pairs obtained from the DiffusionDB data set. We also demonstrated how a model trained on the image-to-text task can be beneficial to the original task, generating images that are more faithful to the input text prompts. 

Learning to predict the original text prompts of diffusion models employed for text-to-image generation is a stepping stone towards gaining a deeper comprehension of these models. Through our research, we open a new path towards the development of more advanced and effective techniques in the area of diffusion modeling.

\section*{Acknowledgments}
This work was supported by a grant of the Ministry of Research, Innovation and Digitization, CCCDI - UEFISCDI, project number PN-IV-P6-6.3-SOL-2024-2-0227, within PNCDI IV.




\section{Ethics statement}

The ethics of reverse engineering depend on various factors, including the context, purpose, and potential consequences of the reverse engineering activity. In our case, reverse engineering the Stable Diffusion model is aimed at gaining several advantages, as highlighted below:
\begin{itemize}
\item Our experiments show how well the Stable Diffusion model can be reversed, indicating the level of misalignment between prompts and images. This allows researchers and practitioners to gain a deep understanding of the behavior of this generative diffusion model, which is crucial for improving model performance, identifying potential weaknesses, and enhancing interpretability of results.
\item Examining the internals of the Stable Diffusion allows researchers to identify sources of errors or limitations, and implement targeted fixes, improving the robustness and reliability of the model. A concrete example of this situation is the application of the fine-tuned U-Net to generate images that are better aligned with the input text.
\item Reverse engineering can lead to the discovery of novel algorithms or improvements to existing ones. This can contribute to the advancement of the broader field of generative models and diffusion models, fostering innovation and pushing the boundaries of what is currently possible.
\item Our study also represents a way to assess the robustness of generative diffusion models to reverse engineering. Understanding potential vulnerabilities is crucial for addressing security concerns and ensuring that models are not exploited for malicious purposes. Additionally, it helps in addressing ethical considerations related to the use of such models.
\end{itemize}

While these benefits can be substantial, it is important to note that our reverse engineering study was conducted ethically and in compliance with legal and ethical standards, ensuring the responsible use of open-sourced technology.


\bibliographystyle{model2-names}
\bibliography{references}

\end{document}